\documentclass[11pt,a4paper]{article}
\usepackage[hyperref]{acl}
\usepackage{times}
\usepackage{latexsym}

\usepackage{microtype}

\aclfinalcopy 

\title{Meeting Summarization with Pre-training and Clustering Methods}
\author{Andras Huebner\\
  \texttt{andreas.huebnerh@gmail.com} \\
  Wei Ji \\
  \texttt{jiwei0706@gmail.com} \\
  Xiang Xiao \\
  \texttt{mitchell.xiao@gmail.com}}
\date{\today}

\begin{document}
\maketitle
\begin{abstract}
Automatic meeting summarization is becoming increasingly popular these days. The ability to automatically summarize meetings and to extract key information could greatly increase the efficiency of our work and life. In this paper, we experiment with different approaches to improve the performance of query-based meeting summarization. We started with HMNet\cite{hmnet}, a hierarchical network that employs both a word-level transformer and a turn-level transformer, as the baseline. We explore the effectiveness of pre-training the model with a large news-summarization dataset. We investigate adding the embeddings of queries as a part of the input vectors for query-based summarization. Furthermore, we experiment with extending the locate-then-summarize approach of QMSum\cite{qmsum} with an intermediate clustering step. Lastly, we compare the performance of our baseline models with BART, a state-of-the-art language model that is effective for summarization. We achieved improved performance by adding query embeddings to the input of the model, by using BART as an alternative language model, and by using clustering methods to extract key information at utterance level before feeding the text into summarization models.

The code of our experiments can be found in our github repository \url{https://github.com/wxj77/MeetingSummarization}.
\end{abstract}

\section{Introduction}

Meeting summarization models aim to extract key information from video records, audio records, or meeting transcripts that consist of utterances from multiple people on multiple topics. Meeting is a common subject in everyday life, and millions of meetings happen in the US and around the world every day. An automatic meeting summarization model can help both meeting participants and non-participants review key information of a meeting in a shorter time and potentially save a significant amount of time by making the usual manual note taking at least partly obsolete.

One essential challenge of meeting summarization is that the discussions in a meeting are usually multi-faceted and different participants may hold different opinions or focus on different facets of the same object, therefore, it is challenging to compress or compose a short summary that contains all the salient information. \cite{qmsum} proposed QMSum, a new meeting summarization pipeline that adopts a query-based summarization approach. QMSum adopted a two-stage locate-then-summarize solution. Given a query, a Locator model first locates the relevant text spans in the meeting according to the query, then a Summarizer model summarizes the selected text spans based on the query. This approach is more flexible and interactive, and caters to users’ diverse intents when generating summaries. In this paper, we aim to build on top of QMSum and explore ways to make it generates better meeting summaries.

As our baseline model, we used the baseline model presented by \cite{qmsum}. In our experiments, since we mainly focus on the Summarizer, we directly used the provided relevant text spans for each query, published by the authors. For the Summarizer, we experimented both with the HMNet model\cite{hmnet} and the BART model\cite{lewis2019BART}. For the HMNet model, we used a simplified version\cite{hmnetsimple} and extended/changed the code as required for our experiments. For the BART model, we used the Fairseq sequence modeling toolkit\cite{ott2019fairseq}, which provides implementation of BART finetuned on the CNN-DailyMail dataset\cite{cnn}. 

We used the provided relevant text spans of the Locator and the simplified version of the HMNet Summarizer without query embedding as the baseline for our experiments.

We performed the following experiments with the assumption that they might help improving the performance of this model:
\begin{itemize}
\item Adding query embedding into the code base of the simplified HMNet
\item Pre-Training the simplified HMNet model with CNN data
\item Pre-Training the simplified HMNet model with CNN data most related domain knowledge to the QMSum dataset
\item Testing a locate-then-cluster-then-summarize approach (while the original HMNet implementation uses a locate-then-summarize approach)
\item Testing an alternative language model (BART) 
\end{itemize}

\section{Related Work}

Meeting summarization models can be grouped into two major categories: extractive methods and abstractive methods. The extractive methods use words picked from the original text to generate summaries and therefore has a higher chance to maintain faithfulness of the original text. However, discussions in meetings are usually multi-faceted and consist different opinions from different people, thus, extractive methods may not capture all key information of a meeting. On the other hand, the abstractive methods use language models and encoding-decoding algorithms, which can generate better meeting summaries. Because abstractive meeting summarization is the mainstream approach and is more commonly discussed in recent papers, we focus on discussing recent abstractive meeting summarization methods. 

\cite{qmsum} present an end-to-end learning network that leverages the encoder-decoder transformer architecture. They introduce two major design improvements to adapt the structure to meeting summarization: 1) employ a two-level hierarchical structure: a word-level transformer to process the token sequence of one turn (i.e., an utterance of a participant) in the meeting, and a turn-level transformer to process the information of all m turns in a meeting; 2) train a role vector for each meeting participant during encoding, and append this role vector to the turn-level representation for later decoding. 

\cite{long_dialogue_summarization} perform a comprehensive study on long dialogue summarization by investigating strategies to address the lengthy input problem. The authors investigate the following models: 1) a retrieve-then-summarize pipeline\cite{graph_model}; 2) end-to-end summarization models that include BART (a transformer-based encoder-decoder model), HMNet (a hierarchical type of model)\cite{hmnet}, and Longformer (an extended transformer model). The authors compare performance of these models in generating summaries for long dialogue and find that the retrieve-summarize pipeline results in the best performance, and BART-large performs worse than HMNet when the input is beyond 512 tokens. 

\cite{multi_sentence_compression} present a fully unsupervised, end-to-end meeting speech summarization framework that composes 4 modules: 1) text preprocessing; 2) cluster utterances that should be summarized by a common abstractive sentence; 3) generate a single abstractive sentence for each utterance community, using an extension of the Multi-Sentence Compression Graph; 4) perform budgeted submodular maximization to select the best elements from the generated sentences to form the summarization. Advantages of the proposed approach are that it is fully unsupervised and can be interactively tested.

\cite{graph_model} explore dialogue discourse (i.e., relations between utterances in a meeting) to model the utterances interactions for meeting summarization. Specifically, the authors first convert meeting utterances with discourse relations (6 types) into a meeting graph where utterance vertices interact with relation vertices. Then, they design a graph-to-sequence framework to generate meeting summaries using a standard LSTM decoder with attention and copy mechanisms.

Besides introducing the QMSum dataset, \cite{qmsum} also introduce a baseline model for query-based meeting summarization. They adopted a two-stage locate-then-summarize solution: first, a Locator model locates the relevant text spans in the meeting according to the queries, and then a Summarizer model summarizes the selected text spans based on the query. The authors investigated variants of the model with different Locators and Summarizers. They found that a Pointer Network and hierarchical ranking-based Locator and the HMNet Summarizer generated the best summarization results. 

\section{Data}

As the basis for our experiments, we use the AMI part of the QMSum dataset and combine it with the extracted relevant spans for each query (published results of the Locator part of the QMSum model) \cite{qmsumspan}. This was done by using the original AMI data. In the original AMI data, each dataset represents one meeting and the dataset contains overall 137 datasets, but for each dataset, there exist multiple combinations of specific queries and reference summaries (answers) for each query. 

For our experiments, we generated for each combination of meeting transcript/query (both for train and test meeting transcripts) a separated dataset in the required input format which resulted in 894 training datasets and 196 test datasets (each dataset representing a unique combination of query/meeting transcript). Each meeting trancript appears multiple times (as each meeting transcript has multiple queries with reference summaries included in the QMSum dataset, but the combination of each query/summarization/meeting transcript is unique).

For experiments including pre-training of the model with news summarization data, we use parts of the CNN news summmarization data (which are part of the CNN/DailyMail dataset) \cite{cnn}. The CNN/DailyMail dataset consists of ~312k news articles with reference summaries, the CNN subpart consists of 92.5k news articles.

\section{Our models}
Motivated by locate-then-summarize method\cite{qmsum} and unsupervised graph-base multi-sentence compression (MSC) models\cite{multi_sentence_compression}, we would like test whether we can combine benefit of the two approaches. We proposed a approach of locate-then-cluster-then-summarize by adding a unsupervised clustering layer between the locator and the summarizer. The added clustering layer can help further extract the key information from each utterance and reduce unrelated short utterances. 

We build our baseline model based on the QMSum pipeline. The original QMSum model consists of two parts: The Locator and the Summarizer. The Locator recommended by the authors is a hierarchical ranking based Locator. The authors also published the manually labeled relevant 'golden' text spans for each pair of  summarized text and meeting script. We decided to use the published relevant text spans as our input.

The authors of the QMSum model tested different models as the summarizer and found HMNet to be the best performing summarizer. Therefore, we used a simplified implementation of the HMNet model \cite{hmnetsimple} as our baseline model. The reason for choosing this simplified implementation was that the original HMNet model requires more GPU resources than we had available.

The simplified version of HMNet consists, as the original implementation, of a Word-level Transformer, followed by the Turn-level Transfomer and a Decoder. In the original implementation, the inputs into the Word-level Transformer are the encoded tokens of each turn, the encoded role of the speaker and two embedding matrices to represent the part-of-speech (POS) and entity (ENT) tags. The simplified version doesn't use POS and entity tags.

For our baseline result, we used the simplified HMnet model with the AMI part of the QMSum dataset with the original hyperparameters and no role vectors and no positional encoding included. The training was performed with 30 epochs.

To get the clustering layer in between the locator and the summarizer, we followed the clustering methods in \cite{multi_sentence_compression}. The authors discussed graph-based method to summarize text, in which a word-graph is first generated and then is used to get the best paths in the graph as the summarized results. We used four methods in the paper: \cite{filippova-2010-multi} in which a word graph is generated by iteratively adding sentences to it and the shortest paths in word graph is used as the result, \cite{boudin-morin-2013-keyphrase} in which a word co-occurence network is built to generate the word-graph, \cite{mehdad-etal-2013-abstractive} in which an entailment word-graph over the sentence communities is built, and  \cite{tixier-etal-2016-graph} in which a graph degeneracy-based approach is used to decompose the word-graph to extract the central nodes that contain key information in the word-graph.

We also explored an alternative language model as the summarizer, BART\cite{lewis2019BART}, which is a de-noising pre-trained model for language generation, translation and comprehension. BART is particularly effective when fine tuned for text generation, and has achieved new state-of-the-art results on many generation tasks, including summarization. We used the BART implementation provided by Fairseq\cite{ott2019fairseq}, and finetuned it on the CNN-Dailymail summarization task\cite{cnn}.

We measured the performance of our models with ROUGE-1, ROUGE-2 and ROUGE-L score \cite{rouge}. Rouge-N score is a recall metric which measures what percentage of all n-grams in the human generated reference summary are captured by the n-grams of the algorithmic generated summary. ROUGE-L measures the overlap of the longest common sequence between the algorithmic generated summary and the human generated reference summary, calculating both the recall (number of words in the reference summary in denominator) and the precision (number of words in generated summary in denominator).

\section{Experiments}
\begin{table*}
\centering
\begin{tabular}{llll}
\hline
\textbf{Model} & \textbf{ROUGE-1 F} & \textbf{ROUGE-2 F} & \textbf{ROUGE-L F}\\
\hline
HMNet/Simp. in original form & 0.251 & 0.048 & 0.225 \\
HMNet/Simp. query emb. & \textbf{0.259} & \textbf{0.054} & \textbf{0.233} \\
HMNet/Simp. 10k pre-training & 0.246 & 0.044 & 0.224 \\
HMNet/Simp. most sim. 15k / 15 epochs pre-training & \textbf{0.254} & \textbf{0.042} & \textbf{0.230} \\
HMNet/Simp. most sim. 5k / 5 epochs pre-training & 0.205 & 0.027 & 0.187 \\
BART & 0.252 & 0.061 & 0.222 \\
\hline
\end{tabular}
\caption{\label{tab: qmsum pretrain}
Performance of the models when generating summaries that corresponds to the queries in the QMSum dataset.
}
\end{table*}

\begin{table*}
\centering
\begin{tabular}{llll}
\hline
\textbf{Model} & \textbf{ROUGE-1 F} & \textbf{ROUGE-2 F} & \textbf{ROUGE-L F}\\
\hline
HMNet/Simp. & 0.251 & 0.048 & 0.225 \\
HMNet/Simp. + Clustering utterances & \textbf{0.263} & \textbf{0.052} & \textbf{0.233} \\
HMNet/Simp. + Longest Sentences & 0.242 & 0.047 & 0.221 \\
BART & 0.252 & 0.061 & 0.222 \\
BART + Clustering utterances & \textbf{0.273} & \textbf{0.067} & \textbf{0.237} \\
BART + Longest Sentences & 0.250 & 0.058 & 0.220 \\
\hline
\end{tabular}
\caption{\label{tab: qmsum cluster}
Performance of the models after clustering.
}
\end{table*}

\begin{table*}
\centering
\begin{tabular}{llll}
\hline
\textbf{Model} & \textbf{ROUGE-1 F} & \textbf{ROUGE-2 F} & \textbf{ROUGE-L F}\\
\hline
BART (summaries for queries) & \textbf{0.252} & \textbf{0.061} & \textbf{0.222} \\
BART (summaries for entire meeting) & 0.227 & 0.047 & 0.205 \\
\hline
\end{tabular}
\caption{\label{tab: BART}
BART performance when generating summaries that corresponds to the queries vs. summaries for the entire meetings. 
}
\end{table*}

\subsection{Baseline model}
We used the code of the simplified HMNet model with QMSum data as input with default settings (no speaker role vectors, no POS tagging) as our baseline. For our first experiment, we embedded the query for each query/meeting summarization dataset as the first utterance of a meeting, using the fictive meeting role "questioner". The hyperparameters of the model remained unchanged over all experiments if not specifically called out otherwise in the following sections.

\subsection{Pre-training with domain data}
We used the model above and pre-training it on 10k randomly selected CNN news from the CNN-DailyMail data set (5 epochs pre-training). To achieve this, we expand the code of the simplified HMNet model to allow transforming and loading the CNN news summarization data into a pseudo-meeting format. Then, we perform a pre-training loop before fine-tuning the model on the QMSum training data. 

To further see whether the size of pre-training data could impact model performance, we used the model above and increased the size of the pre-training news summarization datasets to 15k. We wrote a custom logic to identify the 15k news whose topics are most related to QMSum meetings, and used them for pre-training purposes (in contrast to just choosing news datasets randomly). The pre-training was again performed over 15 epochs. We gain very small improvement by pre-training with 15k related news data. 

In order to further evaluate the effect of pre-training, we ran another experiment to fine-tune the model with only the 5k CNN news that are most related to the AMI meeting topics (5 epochs). However, we observed a decrease in performance. Results are shown in table \ref{tab: qmsum pretrain}. 

\subsection{Clustering utterances}
\label{subsection:Clustering utterances}
Inspired by unsupervised meeting summarization methods \cite{multi_sentence_compression}, we first generated a short meeting script by using four different clustering method in \cite{filippova-2010-multi,boudin-morin-2013-keyphrase,mehdad-etal-2013-abstractive,tixier-etal-2016-graph} to cluster each utterance in given text span. We found some utterances that we cannot find candidate results of clustering. These sentences mostly consists of just stopping words, such as 'Hmm', 'Right'. But we also find some meaningful sentences also yield to no clustering results. We think it could be because the length of the utterance is too short to generate a word graph to be clustered. So we chose to use the longest sentence of that utterance in the short meeting script. After clustering each utterance. we combined the generated short meeting scripts from the four different clustering methods and then used the combined short meeting script as the input and pass it through Simple HMNet to compare with our baseline model. Results are shown in table \ref{tab: qmsum cluster}. We achieved improvement of ROUGE scores by this method. 

\subsection{Longest sentences}
To further explore the impact of reducing the length of the utterance, we decided to generated a short meeting script by only using the longest sentence of each utterance. Results are shown in table \ref{tab: qmsum cluster}. We notice that the ROUGE scores are comparable to that of the baseline model. This finding suggests that a shorter utterances of extracted key information would yield to as good result as long utterances.

\subsection{BART}
We performed the following experiments with the BART model:
\begin{itemize}
\item Using BART to generate summaries that correspond to the queries in the QMSum dataset
\item Using BART to generate summaries for the entire meeting transcripts in the QMSum dataset 
\end{itemize}

We first compared the performance of BART with that of the HMNet models. In \cite{long_dialogue_summarization}, the authors found that the overall performance
of HMNet is worst than BART. Therefore, we expect BART to perform better than the simplified HMNet models explored. Second, we studied whether the performance of BART is affected by the input length. In \cite{long_dialogue_summarization}, the authors found that the performance of the BART model decreases sharply when the dialogue input becomes longer. They argue that this could be the result of BART's unique properties: pre-trained on the datasets with a limited length (i.e., 1,024) and the input has to be truncated to fit the limitation. In the QMSum dataset, the transcripts for the entire meeting are usually very long (30 minutes to 1 hour meeting), while the relevant texts pans for each query are much shorter. We expect to achieve similar results from \cite{long_dialogue_summarization}: BART performs better when generating summaries that correspond to the queries (shorter input text) than generating summaries for the entire meeting (longer input text). 

\section{Analysis}
\subsection{Query Embedding}
We found that having the simplified HMNet model as the baseline, and embedding queries into the model resulted in a slight improvement in the ROUGE scores. Therefore, query embedding helps to improve performance of HMNet.

\subsection{Pre-train with related data}
We found that pre-training the model with randomly selected 10k CNN news didn't improve ROUGE scores. But pre-training the model with the most similar 15k CNN news with 15 epochs resulted in an improvement in the ROUGE scores. However, the performance still didn't beat the baseline. Pre-training the model with the most similar 5k CNN news with 5 epochs, interestingly, led to a significantly drop in ROUGE scores. 

The fact that pre-training with CNN datasets didn't help to improve HMNet performance contradicts to our assumption. This might be caused by that the size of the pre-training datasets was not large enough. Looking at the overall results, it seems that ideally the pre-training would need to be performed with a larger number of news articles but with only a small number of pre-training epochs. When looking at the results of the different pre-training data size/epoch combinations, the best combination was pre-training with the most related 15k CNN news and 15 epochs of pre-training. The next logical step would have been to use, for example, the most related 30k news and experimenting with different numbers of pre-training epochs, which worth further investigation.

Another reason why the pre-training approach didn't work might be that the authors of the original HMNet model took a different approach when setting up the pre-training. They chose groups of each 4 news articles to build an "artifical" meeting and in addition, shuffled the sentences of the news articles. Furthermore, they performed the pre-training over the complete CNN/Daily Mail and also XSum datasets with over 500k news articles.

\subsection{Utterance-level information extraction}
We found a performance improvement by using clustering methods to apply a utterance-level information extraction and then apply the summarization model over the meeting scripts. This finding confirmed that summarizing each utterance first and then summarizing the summarized utterances could be give a better meeting summarization result. The utterance-level summarization can dropping unrelated stopping words from the scripts, and pick the most important information from each utterance, and therefore lead to a better result. 

We tried another utterance-level information extraction by using the longest sentence from each utterance as the input. We found that by simply using the longest sentence from each utterance as the information extraction method yield to a worse result. However, the result we got is still comparable to that using the whole meeting scripts. We think this finding suggested that a large portion of information is contained in the longest sentences of each utterance, and therefore we got comparable results. However,because the key information is not just included in the longest sentence but is contained across the utterance, the overall performance is worse than the result using the whole script. We think that to further confirm our finding, other utterance-level information extraction methods are also worth trying to further evaluate the impact of utterance-level information extraction.  

\subsection{BART as the summarizer}
BART achieved a slightly better performance than the simplified HMNet model, which is consistent as the observations in \cite{long_dialogue_summarization}. When looking at the performance of BART as the input length changes(table \ref{tab: BART}), our finding is also consistent to \cite{long_dialogue_summarization}: the performance
of the BART model is worse at generating summaries for the entire meeting (much longer dialogue input) than generating summaries for text spans that correspond to queries (shorter dialogue input).

\section{Conclusion}
In summary, we achieved small improvement of ROUGE scores by applying a clustering algorithm to generate a word-graph of each utterance, condensing the utterance into a shorter sentence by finding the best-path in the word-graph, and using the shorter sentences as the transcript to be summarized. This finding suggests that by using clustering algorithms to pick up the most important words and phrases in each utterance, the performance of meeting summarization would still be maintained or improved. This finding also suggests that an additional first-stage clustering step over each utterance would help yield better results for summarizing the whole meeting transcript. 

We also found that pre-training the summarization model with related data in the same domain could improve the performance of summarization. But this pre-traing method requires a larger quantity of related data from the same domain to show its effectiveness. 

In addition, the performance of the BART model is affected by the input length, and BART performs better at query-based meeting summarization with QMSum than generating sumamries for the full meeting transcripts. 

Also, despite the ROUGE scores, the generated texts still contain many grammatical mistakes which result in poor readability.

\section*{Acknowledgments}

We would like to thank the staff of XCS224U course.

\bibliographystyle{acl}
\bibliography{acl}

\end{document}